%% file: 00main.tex
\documentclass[12pt]{article}
\usepackage[utf8]{inputenc}
\usepackage[T1]{fontenc}
\usepackage{amssymb}
\usepackage{amsmath}
\usepackage{amsfonts}
\usepackage{stfloats}
\usepackage [autostyle, english = american]{csquotes}
\MakeOuterQuote{"}
\usepackage{subcaption}
\usepackage{textcomp}
\usepackage{blindtext}
\usepackage{graphicx}
\usepackage{lettrine}
\usepackage{lipsum}
\usepackage{float}
\usepackage{multicol}
\usepackage{multirow}
\usepackage{enumitem}
\usepackage{textcomp}
\usepackage{layouts}
\usepackage[font=small,skip=0pt]{caption}
\usepackage{color}
\usepackage{mathtools}
\usepackage{booktabs}

\usepackage{array}
\newcolumntype{L}[1]{>{\raggedright\let\newline\\\arraybackslash\hspace{0pt}}m{#1}}
\newcolumntype{C}[1]{>{\centering\let\newline\\\arraybackslash\hspace{0pt}}m{#1}}
\newcolumntype{R}[1]{>{\raggedleft\let\newline\\\arraybackslash\hspace{0pt}}m{#1}}
\usepackage[linesnumbered]{algorithm2e}
\usepackage{bbm}
\usepackage{duckuments}
\usepackage[hidelinks]{hyperref}
\usepackage[hidelinks]{hyperref}
\usepackage{authblk}
\usepackage{epstopdf}
\usepackage{etoolbox}
\usepackage{authblk}
\usepackage{epstopdf}
\usepackage{etoolbox}
\makeatletter
\def\@seccntformat#1{\@ifundefined{#1@cntformat}%
   {\csname the#1\endcsname.\hskip0.5em}    % default
   {\csname #1@cntformat\endcsname}% enable individual control
}
\appto{\appendix}{%
    \renewcommand{\appendixname}{Appendix}
    \newcommand{\section@cntformat}{\appendixname~\Alph{section}.\hskip0.5em}
    \newcommand{\subsection@cntformat}{\Alph{section}.\arabic{subsection}.\hskip0.5em}}
\makeatother

\begin{document}
% \begin{frontmatter}

\title{Nonlinear Model Predictive Control of Tiltrotor Quadrotors with Feasible Control Allocation}

\author[1]{Zeinab Shayan}
\author[1]{Jann Cristobal}
\author[1]{Mohammadreza Izadi}
\author[1]{Amin Yazdanshenas}
\author[2]{Mehdi Naderi}
\author[1,*]{Reza Faieghi}
\affil[1]{\small{Autonomous Vehicles Laboratory, Department of Aerospace Engineering, Toronto Metropolitan University, 350 Victoria St., M5B2K3, Toronto, Ontario, Canada}}
\affil[2]{\small{School of Production Engineering and Management, Technical University of Crete, Chania, Greece}}
\affil[*]{Corresponding author: Email: reza.faieghi@torontomu.ca}
\date{}
\maketitle
\begin{abstract}
This paper presents a new flight control framework for tilt-rotor multirotor uncrewed aerial vehicles (MRUAVs).
Tiltrotor designs offer full actuation but introduce complexity in control allocation due to actuator redundancy.
We propose a new approach where the allocator is tightly coupled with the controller, ensuring that the control signals generated by the controller are feasible within the vehicle actuation space.
We leverage nonlinear model predictive control (NMPC) to implement the above framework, providing feasible control signals and optimizing performance. 
This unified control structure simultaneously manages both position and attitude, which eliminates the need for cascaded position and attitude control loops.
Extensive numerical experiments demonstrate that our approach significantly outperforms conventional techniques that are based on linear quadratic regulator (LQR) and sliding mode control (SMC), especially in high-acceleration trajectories and disturbance rejection scenarios, making the proposed approach a viable option for enhanced control precision and robustness, particularly in challenging missions.

% %%Graphical abstract
% \begin{graphicalabstract}
% \includegraphics{grabs}
% \end{graphicalabstract}

%%Research highlights
% \begin{highlights}
% \item A tightly coupled controller-allocator framework for tiltrotor MRUAVs that ensures feasible control inputs.
% \item Nonlinear model predictive flight controller for tiltrotor MRUAVs, for the first time.
% \item Eliminating the cascaded control architecture for tiltrotor MRUAVs, simultaneously optimizing position and attitude control.
% \end{highlights}

\textbf{Keywords:}
Uncrewed aerial vehicles, quadrotor, omnidirectional quadrotor, nonlinear model predictive control, control allocation
%% keywords here, in the form: keyword \sep keyword
\end{abstract}

\input{01Intro}
\input{02Modeling}
\input{03Controller}

\input{04Results}

\input{05Conclusion}

\bibliographystyle{elsarticle-num} 
\bibliography{References}

\end{document}

%% file: 01Intro.tex
\section{Introduction}\label{se:intro}
In conventional quadrotors and similar multirotor uncrewed aerial vehicles (MRUAVs), all rotors are configured in the same or parallel planes, making them structurally simple \cite{maaruf2022survey,amin2016review,izadi2024high}; however, this results in under-actuation. 
Each vehicle has six degrees of freedom (DOFs) to be controlled, but its control inputs are limited to the moments around the three axes of the body frame, and the thrust that is only alongside the vertical axis. 
As a result, the longitudinal and lateral dynamics are coupled with rotational dynamics, making it difficult to perform maneuvers that require the vehicle to control its position or orientation independently from one another.

Tiltrotor MRUAVs address the above challenge by incorporating extra actuation mechanisms that dynamically tilt the rotors' planes, allowing the vehicle to independently control six DOFs.
Such vehicles can hover at a fixed position while maintaining a constant non-zero attitude, a capability that conventional MRUAVs cannot achieve \cite{lv2021design,hao2021fault,hamandi2021design,yu2022over,marcellini2023px4}.
Tiltrotor MRUAVs are divided into two categories: single-axis \cite{ryll2015novel,kamel2018voliro} and dual-axis \cite{de2017design,bin2018design,von2018dual} tiltrotor.
In the single-axis setup, each rotor tilts around only one axis of the body frame, while in the dual-axis, rotors tilt around two axes.

% \begin{figure}[t]
%     \centering
%     \includegraphics[width=\linewidth]{Fig/eyecatcher.png}
%     \caption{Single-axis tiltrotor quadrotor conducting hovering with non-zero pitch and roll angles}
%     \label{fig:enter-label}
% \end{figure}

The addition of tilting mechanisms often leads to redundancy in MRUAV actuators.
Therefore, an important question that arises is how to optimally distribute the desired control effort among the actuators, a topic that is widely studied in the field of control allocation \cite{johansen2013control} where the control system typically consists of a controller and an allocator.
The controller generates commands, otherwise known as virtual control inputs, for the allocator.
Next, the allocator translates the virtual control inputs into specific commands for each actuator, attempting to construct the desired control effort prescribed by the controller.

A particular challenge to this setup is that the virtual control inputs computed by the controller may not be feasible for the actuators. 
Our strategy to address this is to tightly couple the controller and allocator, incorporating the actuator constraints directly in the controller formulation, and generating virtual control inputs that are feasible for the actuators.

Driven by the above factors, we aim to design a tightly coupled controller-allocator for flight control of tiltrotor MRUAVs.
The developed algorithm generates control signals that are feasible in the actuation space of the vehicle, optimizing the distribution of control efforts among the actuators, and ultimately, enhancing the maneuverability of the vehicle.
In our approach, we leverage nonlinear model predictive control (NMPC) formulation, as it allows us to account for the actuator constraints explicitly.
Despite the full-actuation of tiltrotor MRUAVs, many existing methods still rely on the separation of position and attitude control, using cascaded control architectures.
In the proposed strategy, instead of using the cascaded architecture, we solve 6-DOF flight control in one NMPC computing frame.
This enables optimizing for position and attitude control at the same time, while simplifying the control architecture, and potentially reducing calibration efforts and computational load.
The key features of our method include:
\begin{itemize}[leftmargin=*]
\item A tightly coupled controller-allocator framework for tiltrotor MRUAVs that ensures feasible control inputs.
\item NMPC-based flight controller for tiltrotor MRUAVs, for the first time.
\item Eliminating the cascaded control architecture for tiltrotor MRUAVs, simultaneously optimizing position and attitude control.
\end{itemize}
The above features enable more efficient and effective flight control for tiltrotor MRUAVs compared to the existing results, as will be shown in our comparative studies in Section \ref{se:results}.

\section{Related Work}
% The two important parts of tiltrotor MRUAV flight controllers are the controller and allocator. In this section, we first review the existing approaches to design the controller and then discuss the control allocation algorithms.

\subsection{Flight Controllers for Tiltrotor MRUAVs}
Flight control of tiltrotor MRUAVs has gained increasing interest in recent years.
Notable works include linear quadratic regulator (LQR) with an integral action controller \cite{allenspach2020design} and linear model predictive control (LMPC) \cite{eskandarpour2023constrained}.
The latter intends to design a generic control framework for conventional and tiltrotor MRUAVs; thus, proposes a cascaded control structure that is not necessary for fully actuated MRUAVs.
The methods developed in \cite{allenspach2020design, eskandarpour2023constrained} provide optimal performance in simple maneuvers.
However, since they rely on a linearized model of the vehicle, they struggle in aggressive maneuvers or in the presence of significant external disturbances, as the linear model may not accurately capture the vehicle dynamics in such scenarios.

Furthermore, several studies have explored nonlinear control methods. 
Sliding mode control (SMC) has been a popular choice, perhaps due to its simplicity and robustness, being applied for controlling tiltrotor quadrotors \cite{sridhar2022nonlinear, mpanza2021control} and trirotors \cite{yu2021immersion, yu2023thrust}.
Other nonlinear methods include feedback linearization \cite{saif2017feedback,ryll2015novel}, adaptive control \cite{csenkul2014adaptive, ding2020tilting}, and backstepping \cite{saif2018decentralized}.
The main drawback of these approaches is the loose coupling between the controller and the allocator.
The controller can generate virtual control inputs that are infeasible for the actuators. 
This leads to unwanted actuator saturation which may lead to performance degradation, as will be shown in our numerical experiments, or even instability in severe conditions.

In addition to these theoretical advancements, significant progress has been made in enabling tiltrotor MRUAV control with open-source autopilot programs. 
For instance, in \cite{marcellini2023px4, DAngelo2024}, a modified version of the PX4 Autopilot was developed to implement flight control strategies on real-world single-axis tiltrotor quadrotors.

Of note, despite the recent advances in NMPC for MARUV control \cite{sun2022comparative, nan2022nonlinear, wang2021efficient, carlos2020efficient}, we did not find a study that has explored the application of NMPC for tiltrotor MRUAVs. 
Our NMPC implementation here addresses the drawbacks of the aforementioned linear optimal and nonlinear methods, at the cost of computational load.
However, using efficient solvers such as ACADOS \cite{verschueren2022acados} alongside our relatively simple control architecture ensures real-time performance for our NMPC approach.

\subsection{Control Allocation}
Control allocation is an important part of flight controllers for tiltrotor MRUAVs.
A comprehensive survey of primary control allocation algorithms is available in \cite{johansen2013control}.
The simplest algorithm is the pseudo-inverse method.
This involves building a control effectiveness matrix that maps control inputs to virtual control inputs and computing the pseudo-inverse of that matrix to find the inverse map needed for control allocation \cite{johansen2013control}.
This method has proven effective in many applications, especially when the virtual control input is feasible within the system actuation space \cite{naderi2019guaranteed, naderi2019fault}.
However, if the control effectiveness matrix is ill-conditioned or singular, the pseudo-inverse may not provide a reliable solution.
Moreover, the pseudo-inverse on its own does not incorporate a mechanism to monitor or prevent actuator saturation. 
There exist iterative and constrained optimization methods that address the above concerns, e.g., redistributed pseudo-inverse \cite{jin2005modified}, daisy chain \cite{buffington1996lyapunov}, direct allocation \cite{durham1993constrained}, linear programming \cite{bodson2002evaluation}, and quadratic programming \cite{harkegard2002efficient}.

Existing studies in tiltrotor MRUAV control have used the above control allocation algorithms interchangeably.
In this work, since we ensure the virtual control inputs are feasible within the vehicle actuation space, we use the simple pseudo-inverse approach; however, it can be readily switched with more sophisticated algorithms.

%% file: 02Modeling.tex
\section{Modeling}\label{se:modeling}
% Note that throughout the paper, unless otherwise is mentioned, scalars are in italics, vectors in lowercase bold, and matrices in uppercase bold.

The key difference between the flight dynamics of tiltrotor MRUAVs and conventional ones is how propulsive forces and moments are generated. 
Since our focus is on single-axis tiltrotor quadrotors, we present the flight dynamics model for this particular class of tiltrotor MRUAVs; however, our modeling approach can be generalized to dual-axes tiltrotor MRUAVs with an arbitrary number of rotors.

\begin{figure}[t]
    \centering
        \includegraphics[trim={8.5cm, 4.5cm, 9cm, 2cm}, clip, width = \linewidth]{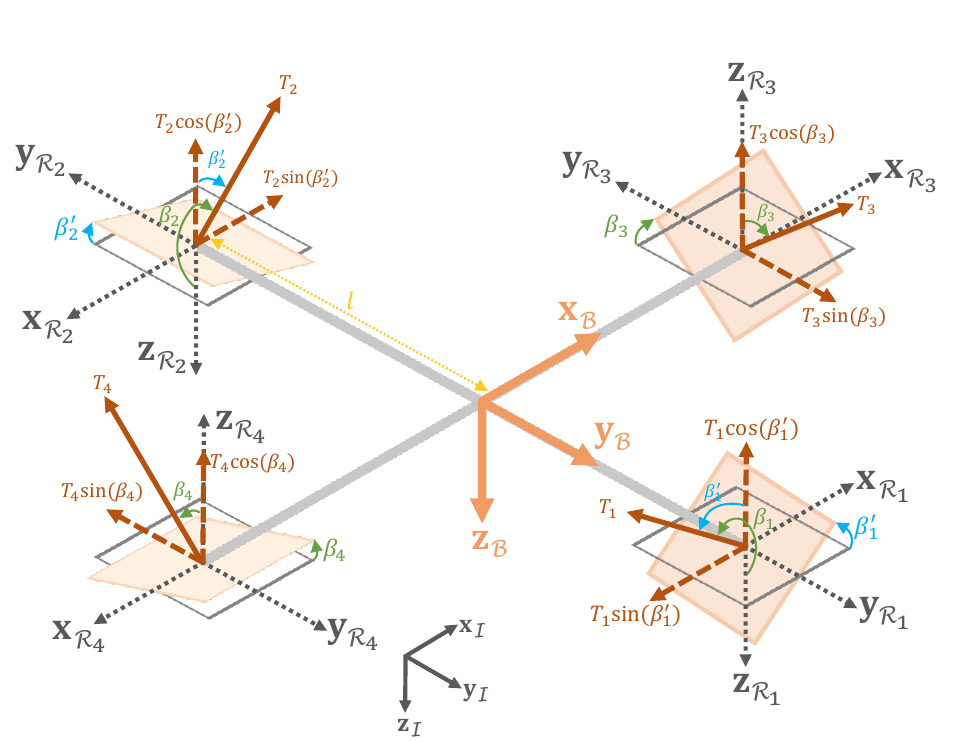}
    \caption{Coordinates frames and parameters setup in our modeling approach for single-axis tiltrotor quadrotor}
    \label{fig:modeling}
\end{figure}

Let $\mathcal{I}=\left\{{\bf{x}}_\mathcal{I},{\bf{y}}_\mathcal{I},{\bf{z}}_\mathcal{I} \right\}$ denote the inertial coordinates frame, $\mathcal{B}=\left\{{\bf{x}}_\mathcal{B},{\bf{y}}_\mathcal{B},{\bf{z}}_\mathcal{B} \right\}$ the body frame, as shown in Fig. \ref{fig:modeling}.
Let ${\boldsymbol{\xi}}=\left[x,y,z\right]^T$ denote the vehicle position, and $\boldsymbol{\eta}  = {[\phi,\theta,\varphi ]^T}$ denote the attitude, where $ - \pi  < \phi  \le \pi $, $ - \frac{\pi }{2} \le \theta  \le \frac{\pi }{2}$, and $ - \pi  < \psi  \le \pi $ are the Euler angles representing roll, pitch, and yaw in the yaw-pitch-roll sequence. Let $\boldsymbol{\upsilon}=\left[u,v,w\right]^T$ denote the linear velocity, ${\boldsymbol{\omega}}=\left[p,q,r\right]^T$ the angular velocity, and ${_\mathcal{I}}{\bf{R}}_{\mathcal{B}}$ the rotation matrix from $\mathcal{B}$ to $\mathcal{I}$.

The transnational dynamics take the following form
\begin{equation}
\label{eq:translationalDyn}
\ddot{\boldsymbol{\xi}} = {\bf{g}} + \frac{1}{m}\left({_\mathcal{I}}{\bf{R}}_{\mathcal{B}}{\bf{f}}_P^\mathcal{B}-{\bf{A}}_T{\boldsymbol{\upsilon}}+{\bf{f}}_D\right),
\end{equation}
where ${\bf{g}}=[0,0,9.81]^T\;m/s^2$ is the gravity vector, $m$ is the vehicle mass, ${\bf{A}}_T$ is the transnational drag coefficient, and ${\bf{f}}_P^\mathcal{B}$ and ${\bf{f}}_D$ represent propulsive and disturbance forces.
The superscript $\mathcal{B}$ is to highlight ${\bf{f}}_P$ is expressed in $\mathcal{B}$. 
This will become useful in the formulation of the propulsive forces and moments to be detailed shortly.

The rotational dynamics of the vehicle take the following form
\begin{equation}
\label{eq:rotationalDyn}
\begin{array}{c}
     \dot{\boldsymbol{\eta}}={\bf{H}}\left(\boldsymbol{\eta}\right)\boldsymbol{\omega},  \\
     {\boldsymbol{\dot \omega }} = {{\bf{J}}^{-1}}\left(-\boldsymbol{\omega} \times {\bf{J}}{{\boldsymbol{\omega }} + {{\boldsymbol{\tau}}_P^{\cal B}} - \bf{A}}_R\boldsymbol{\omega} +{\boldsymbol{\tau}}_d\right),
\end{array}
\end{equation}
where $ \times $ indicates the cross product, ${\bf{J}}$ is the vehicle inertia matrix, ${\bf{A}}_R$ is the rotational drag coefficient, ${\boldsymbol{\tau}}_P^\mathcal{B}$ and ${\boldsymbol{\tau}}_D$ represent propulsive and disturbance moments, and
\begin{equation}
\label{eq:H}
{\bf{H}}\left( {\boldsymbol{\eta }} \right) = \left[ {\begin{array}{*{20}{c}}
1&{\sin \phi \tan \theta }&{\cos \phi \tan \theta }\\
0&{\cos \phi }&{ - \sin \phi }\\
0&{{{\sin \phi } \mathord{\left/
 {\vphantom {{\sin \phi } {\cos \theta }}} \right.
 \kern-\nulldelimiterspace} {\cos \theta }}}&{{{\cos \phi } \mathord{\left/
 {\vphantom {{\cos \phi } {\cos \theta }}} \right.
 \kern-\nulldelimiterspace} {\cos \theta }}}
\end{array}} \right].
\end{equation}

To formulate the propulsive forces and moments ${\bf{f}}_P$ and ${\boldsymbol{\tau}}_P$, we start by defining the $i$-th rotor coordinates frames $\mathcal{R}_i$ with the following orientation with respect to $\mathcal{B}$, also illustrated in Fig. \ref{fig:modeling},
\begin{equation}
\label{eq:rotorFramesSetup}
\begin{array}{l}
{_\mathcal{B}}{\bf{R}}_{\mathcal{R}_1} = \operatorname{diag}\left( 1, 1, 1 \right),\\
{_\mathcal{B}}{\bf{R}}_{\mathcal{R}_2} = \operatorname{diag}\left(-1,-1, 1 \right),\\
{_\mathcal{B}}{\bf{R}}_{\mathcal{R}_3} = \operatorname{diag}\left( 1,-1,-1 \right),\\
{_\mathcal{B}}{\bf{R}}_{\mathcal{R}_4} = \operatorname{diag}\left(-1,1,-1 \right).
\end{array}
\end{equation}
Next, we represent the thrust, torque, and tilting angle of the $i$-th rotor by $T_i$, $Q_i$, and $\beta_i$. 
We note that
\begin{equation}
\label{eq:thrust}
    T_i \approx k_T \Omega_i^2, \; Q_i \approx k_Q \Omega_i^2,
\end{equation}
where $k_T$ and $k_Q$ are rotor coefficients, and $\Omega_i$ is the angular velocity of rotor with opposite signs $\Omega_{1,2} > 0$ and $\Omega_{3,4} < 0$. 
The propulsive force due to $T_i$, denoted by ${\bf{f}}_{P_i}$ is expressed in $\mathcal{R}_i$ as follows
\begin{equation}
\label{eq:rotorForces}
\begin{array}{ll}
  {\bf{f}}_{P_i}^{\mathcal{R}_i} = \left[T_i\sin\left({\beta_i}\right), 0,T_i \cos\left(\beta_i\right) \right]^T,   & i = 1,2, \\
  {\bf{f}}_{P_i}^{\mathcal{R}_i} = \left[0, -T_i\sin\left(\beta_i\right),T_i\cos\left(\beta_i\right) \right]^T,   & i = 3,4.
\end{array}
\end{equation}
Therefore, the total propulsive force expressed in $\mathcal{B}$ is
\begin{equation}
\label{eq:propulsiveForces}
{\bf{f}}_P^{\mathcal{B}} = \sum_{i=1}^4{_\mathcal{B}}{\bf{R}}_{\mathcal{R}_i}{\bf{f}}_{i}^{\mathcal{B}}.
\end{equation}

To construct $\boldsymbol{\tau}_P^\mathcal{B}$, let $l$ represent the distance of rotor from the vehicle centre of mass. 
Then, the moments due to the rotors' thrusts and torques take the following form 
\begin{equation}
\label{eq:propulsiveMoments}
\begin{array}{lll}
    \tau_{P_x}^{\mathcal{B}} & = & {l}\left( f_{1_z}^\mathcal{B} - f_{2_z}^\mathcal{B}\right), \\
    \tau_{P_y}^{\mathcal{B}} & = & {l}\left(-f_{3_z}^\mathcal{B} + f_{4_z}^\mathcal{B}\right), \\
\tau_{P_z}^{\mathcal{B}} & = & \frac{k_Q}{k_T}\left(-f_{1_z}^\mathcal{B} - f_{2_z}^\mathcal{B} + f_{3_z}^\mathcal{B} + f_{4_z}^\mathcal{B}\right) \\ 
&   & + {l}\left( - f_{1_x}^\mathcal{B} + f_{2_x}^\mathcal{B}\right) \\ 
&   & + {l}\left( f_{3_y}^\mathcal{B} - f_{4_y}^\mathcal{B}\right).

\end{array}
\end{equation}

% \begin{equation}
% \label{eq:propulsiveMoments}
% {{\boldsymbol{\tau}}_P^{\mathcal{B}}} = \left[
% \begin{array}{c}
% {l}\left(f_{1_z}^\mathcal{B} - f_{2_z}^\mathcal{B}\right)\\
% {l}\left( -f_{3_z}^\mathcal{B} + f_{4_z}^\mathcal{B}\right)\\
% \frac{k_Q}{k_T}\left(\left(-f_{1_z}^\mathcal{B} - f_{2_z}^\mathcal{B} + f_{3_z}^\mathcal{B} + f_{4_z}^\mathcal{B}\right) + {l}\left( - f_{1_x}^\mathcal{B} + f_{2_x}^\mathcal{B}\right) + {l}\left( f_{3_y}^\mathcal{B} - f_{4_y}^\mathcal{B}\right)\right)
% \end{array}
% \right].
% \end{equation}

Having established ${\bf{f}}_P$ and ${\boldsymbol{\tau}}_P$, it is easy to verify that when $T_i \neq 0$ and $\beta_i \neq 0$, the vehicle can exert forces and moments in all its DOFs. 
This eliminates the underactuation of conventional quadrotors.
Therefore, it is no longer necessary to control position and attitude in a cascaded architecture, a feature that we will leverage in our control design in the next section.

It is noteworthy that our systematic approach in constructing the propulsive forces and moments can be easily extended to other vehicle configurations with an arbitrary number of rotors and single-axis or dual-axes rotor tilting mechanisms.

%% file: 03Controller.tex
\section{Flight Control}\label{se:control}
\begin{figure}[t]
    \centering
    \includegraphics[trim={11cm, 4.5cm, 2cm, 3.3cm}, clip, width = \linewidth]{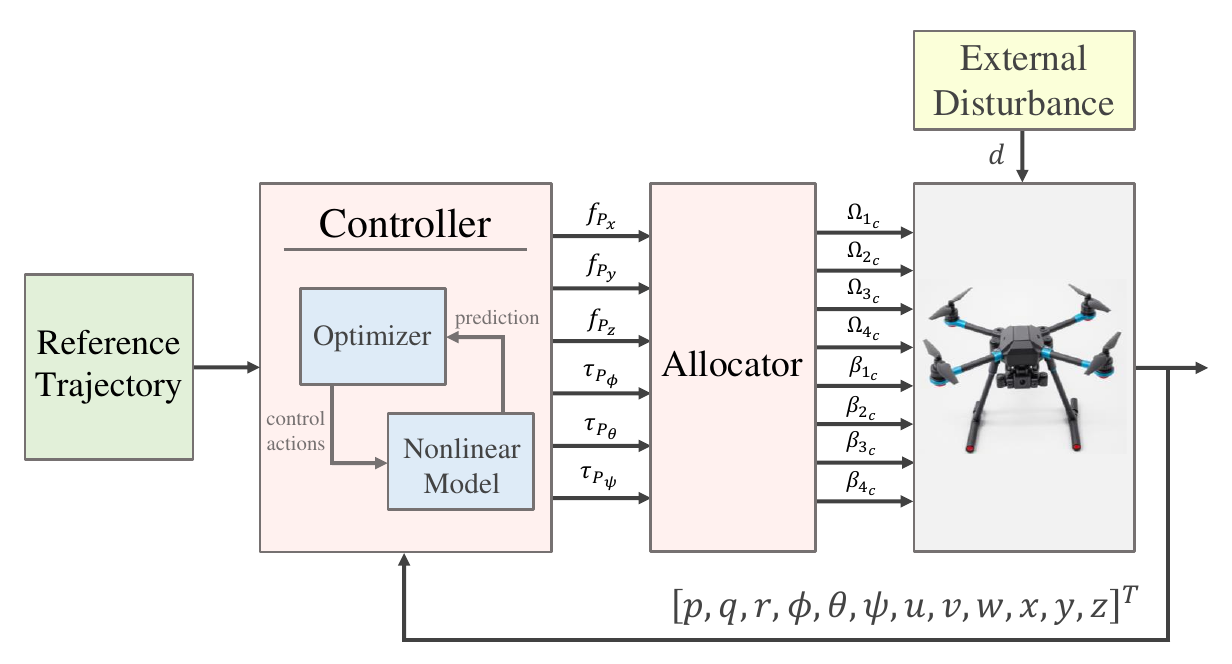}
    \caption{Overview of the proposed flight control framework for single-axis tiltrotor quadrotor}
    \label{fig:blockDiagram}
\end{figure}

Figure \ref{fig:blockDiagram} presents an overview of the proposed flight control framework.
As mentioned earlier, a key feature in our design is to eliminate the cascaded position and attitude control loops,  capitalizing on the extended actuation space of tiltrotor MRUAVs to simplify the control system, and optimize for position and attitude control at the same time.
Therefore, at each computing frame, the controller generates ${\bf{f}}_{P_c}$ and ${\boldsymbol{\tau}}_{P_c}$, collectively called virtual control signals $\mathbf{v} = \left[{\bf{f}}_{P_c}^T, {\boldsymbol{\tau}}_{P_c}^T\right]^T \in \mathbb{R}^6$. 
The subscript $c$ indicates command signals.

The vehicle actuators include four rotors and four servomotors for tilting the rotors.
Therefore, a control allocation algorithm needs to map the virtual control input ${\bf{v}} \in \mathbb{R}^6$ to control input 
\begin{equation}\label{eq:uvector}
    {\bf{u}} = \left[ \Omega_{1_c}, \Omega_{2_c}, \Omega_{3_c}, \Omega_{4_c}, \beta_{1_c}, \beta_{2_c}, \beta_{3_c}, \beta_{4_c} \right]^T \in \mathbb{R}^8.
\end{equation}

This is where the second key feature of our design comes to play. 
In many existing designs, it is not guaranteed that the virtual control signals, i.e. ${\bf{v}}$, generated by the controller leads to a control signal vector {\bf{u}} that is feasible for the actuators.
To ensure ${\bf{v}}$ is feasible within the vehicle actuation space, we tightly couple the controller and allocator by accounting for the physical constraints of actuators in the controller design.
One suitable control framework to achieve this is NMPC.

However, our NMPC formulation is different than existing NMPC formulations for quadrotors \cite{eskandarpour2023constrained, wang2021efficient, carlos2020efficient}.
The existing approaches apply MPC in cascaded control architecture, often only for position control or attitude control; however, our NMPC approach controls the vehicle position and attitude at the same time.
Having only one controller for both position and attitude control presents several advantages inclusive of simplified control architecture, lower computational load, easier calibration, and also simultaneous optimization of position and attitude control. 

% The remainder of this section provides the details of the controller and the allocator.

\subsection{Controller}
Let us start with defining the state vector ${\bf x} = \left[{\boldsymbol{\xi}}^T,{\boldsymbol{\eta}}^T,{\boldsymbol{\upsilon}}^T,{\boldsymbol{\omega}}^T\right]^T$.
The goal is to design the virtual control signal ${\bf v} = \left[{\bf f}_P^T, {\boldsymbol{\tau}}_P^T\right]^T$ such that ${\bf x}$ reaches a desired trajectory ${\bf x}_d$.
To this end, we adopt the standard NMPC formulation 
\begin{equation}
\label{eq:nmpcCost}
  \min \sum_{i=k}^{k+N-1} \ell\left(\mathbf{x}\left(i\right), \mathbf{v}\left(i\right)\right) + \Phi\left(\mathbf{x}\left(k+N\right)\right)
\end{equation}
% \begin{equation}
% \label{eq:nmpcCost}
%   \min_{\{\mathbf{v}\left(k\right), \mathbf{v}\left(k+1\right), \ldots, \mathbf{v}\left(k+N-1\right)\}} \sum_{i=k}^{k+N-1} \ell\left(\mathbf{x}\left(i\right), \mathbf{v}\left(i\right)\right) + \Phi\left(\mathbf{x}\left(k+N\right)\right)
% \end{equation}
subject to
\begin{equation}
\label{eq:nmpcConstraints}
\begin{array}{c}
\mathbf{x}\left(0\right) = \mathbf{x}\left(k\right), \; \mathbf{x}\left(i+1\right) = f\left(\mathbf{x}\left(i\right),\mathbf{v}\left(i\right)\right), \\
\mathbf{x}\left(i\right) \in \mathcal{X}, \; \mathbf{v}\left(i\right) \in \mathcal{V}, \; \forall i \in \{k, \ldots, k+N-1\},
\end{array}
\end{equation}
where $\ell\left(\mathbf{x}\left(i\right), \mathbf{v}\left(i\right)\right)$ is the running cost, $\Phi\left(\mathbf{x}\left(k+N\right)\right)$ is the terminal cost, $\mathbf{x}\left(i+1\right) = f\left(\mathbf{x}\left(i\right), \mathbf{v}\left(i\right)\right)$ is the discretized transnational and rotational dynamics \eqref{eq:translationalDyn} and \eqref{eq:rotationalDyn}, $\mathcal{X}$ and $\mathcal{V}$ are the sets of allowable $\bf{x}$ and ${\bf{v}}$, and $N$ is the prediction horizon.

For our implementations, we set
\begin{equation}
\begin{array}{ccc}
\ell(\mathbf{x}, \mathbf{v}) & = & \left({\bf x} - {\bf x}_d\right)^T {\bf Q_x} \left({\bf x} - {\bf x}_d\right)+ {\bf{v}}^T {\bf Q}_{{\bf{v}}} {{\bf{v}}},
\end{array}
\end{equation}
where $ {\bf Q_x} \ge 0$ and ${\bf Q}_{{\bf{v}}} > 0$, and $\Phi(\mathbf{x}(k+N)) = 0$.

A distinct feature of our approach is how the constraint $\mathbf{v}\left(i\right) \in \mathcal{V}$ is set up.
Instead of explicitly imposing constraints on $\mathbf{v}$, we map $\mathbf{v}$ into $\mathbf{u}$ and use the physical constraints of actuators to find the admissible $\mathbf{u}$. 
This ensures that $\mathbf{v}$ computed from \eqref{eq:nmpcCost} and \eqref{eq:nmpcConstraints} is feasible for the allocator.

To elaborate, let ${\bf{u}}_{\min}$ and ${\bf{u}}_{\max}$ represent the lower and uppper bounds of admissible ${\bf{u}}$.
At each sample time $k$, we use the current state of actuators and their rate of change to update ${\bf{u}}_{\min}$ and ${\bf{u}}_{\max}$, and set
\begin{equation}
\label{eq:constraintOnU}
{\bf{u}}_{\min}\left(k\right) \le h\left(\mathbf{v}\right) \le {\bf{u}}_{\max} \left(k\right),
\end{equation}
where $h\left(\cdot\right)$ is a nonlinear function that maps the virtual control input ${\bf{v}}$ to control input $\bf{u}$.
The development of $h\left(\cdot\right)$ roots into the allocator discussed below.

\subsection{Allocator}
The allocator receives the virtual control input ${\bf{v}} = \left[{\bf{f}}_{P_c}^T, {\boldsymbol{\tau}}_{P_c}^T\right]^T \in \mathbb{R}^6$ from the controller and computes the control input ${\bf{u}}$ in \eqref{eq:uvector}. 
% \begin{equation*}
%     {\bf{u}} = \left[ \Omega_{1_c}, \Omega_{2_c}, \Omega_{3_c}, \Omega_{4_c}, \beta_{1_c}, \beta_{2_c}, \beta_{3_c}, \beta_{4_c} \right]^T \in \mathbb{R}^8. 
% \end{equation*}

To this end, let us probe the propulsive forces of each rotor expressed in \eqref{eq:rotorForces}.
It follows that $\mathbf{f}_{P_{i_y}} = 0$ for $i=1,2$, and $\mathbf{f}_{P_{i_x}} = 0$ for $i=3,4$.
Therefore, the remaining components of propulsive forces that can be used for control allocation is
\begin{equation}
\mathbf{u}'=\left[f_{P_{1_x}},f_{P_{1_z}},f_{P_{2_x}},f_{P_{2_z}},f_{P_{3_y}},f_{P_{3_z}}, f_{P_{4_y}},f_{P_{4_z}}\right]^T.
\end{equation}
Using \eqref{eq:propulsiveForces} and \eqref{eq:propulsiveMoments}, we can write 
\begin{equation}
\label{eq:u_v}
{\bf{v}} = {\bf{B}} \mathbf{u}',
\end{equation}
where 
\begin{equation}
\label{eq:B}
{\bf{B}} = \left[ {\begin{array}{*{20}{c}}
1&0&1&0&0&0&0&0\\
0&0&0&0&1&0&1&0\\
0&1&0&1&0&1&0&1\\
0&{ l}&0&{-l}&0&0&0&0\\
0&0&0&0&0&{ - l}&0&l\\
{-l}&\frac{-k_Q}{k_T}&{ l}&\frac{-k_Q}{k_T}&{ l}&\frac{k_Q}{k_T}&{-l}&\frac{k_Q}{k_T}
\end{array}} \right].
\end{equation}
Multiplying both sides of \eqref{eq:u_v} with $\left({\bf{B}}^T{\bf{B}}\right)^{-1}$ yields
\begin{equation}
\label{eq:uprime}
    \mathbf{u}' = {\bf{B}}^\dag {\bf{v}},
\end{equation}
where ${\bf{B}}^\dag$ is pseudo-inverse of $\bf{B}$.

Once $\mathbf{u}'$ is computed, $\Omega_{i_c}$ and $\beta_{i_c}$ can be obtained from \eqref{eq:thrust} and \eqref{eq:rotorForces} as follows
\begin{equation}
\label{eq:u}
\begin{array}{ll}
\Omega_{1_c} = \frac{30}{\pi}\sqrt{\frac{\left|\mathbf{f}_{P_{1}}\right|}{k_T}}, &
{\beta_{1_c}} =  \arctan\left( \frac{f_{P_{1_x}}}{f_{P_{1_z}}} \right), \\

\Omega_{2_c} = \frac{30}{\pi}\sqrt{\frac{\left|\mathbf{f}_{P_{2}}\right|}{k_T}}, &
{\beta_{2_c}} =  \arctan\left(\frac{f_{P_{2_x}}}{f_{P_{2_z}}} \right),\\

\Omega_{3_c} = -\frac{30}{\pi}\sqrt{\frac{\left|\mathbf{f}_{P_{3}}\right|}{k_T}}, &
{\beta_{3_c}} =  -\arctan\left(\frac{f_{P_{3_y}}}{f_{P_{3_z}}} \right),\\

\Omega_{4_c} = -\frac{30}{\pi}\sqrt{\frac{\left|\mathbf{f}_{P_{4}}\right|}{k_T}}, &
{\beta_{4_c}} =  -\arctan\left(\frac{f_{P_{4_y}}}{f_{P_{_4z}}} \right),

% \Omega_{1_c} = \frac{30}{\pi}\sqrt{\frac{\sqrt{ \left({f_{1_x}^{{\cal R}_1}}\right)^2 + \left({f_{1_z}^{{\cal R}_1}}\right)^2 }}{k_T}},&

% \Omega_{2_c} = \frac{30}{\pi}\sqrt{\frac{\sqrt{ \left({f_{2_x}^{{\cal R}_2}}\right)^2 + \left({f_{2_z}^{{\cal R}_2}}\right)^2 }}{k_T}},\\

% \Omega_{3_c} = -\frac{30}{\pi}\sqrt{\frac{\sqrt{ \left({f_{3_y}^{{\cal R}_3}}\right)^2 + \left({f_{3_z}^{{\cal R}_3}}\right)^2 }}{k_T}},&

% \Omega_{4_c} = -\frac{30}{\pi}\sqrt{\frac{\sqrt{ \left({f_{4_y}^{{\cal R}_4}}\right)^2 + \left({f_{4_z}^{{\cal R}_4}}\right)^2 }}{k_T}}.

\end{array}
\end{equation}
where $\left| \cdot \right|$ indicates the vector magnitude.
Computing \eqref{eq:u} will constitute the control input $\mathbf{u}$ which can be applied to the vehicle.
At the same time, the combination of \eqref{eq:uprime} and \eqref{eq:u} forms $h\left( \cdot \right)$ to be used in the constraint \eqref{eq:constraintOnU} of the NMPC computations, ensuring $\mathbf{v}$ that is fed to the allocator leads to feasible $\mathbf{u}$.

%% file: 04Results.tex
\section{Results}\label{se:results}
This section presents numerical experiments to evaluate the effectiveness of the proposed flight control algorithm.
We compare our method with linear quadratic regulator (LQR) and sliding mode control (SMC), both carefully fine-tuned to achieve their best performance.
Note that both methods control the position and attitude in the same control loop.

The vehicle parameters include $m = 0.468[kg]$, ${\bf{J}} = {\rm{diag}}\left( {4.856,4.856,8.801} \right) \times {10^{ - 3}}[kg{m^2}] $, $l = 0.225[m]$, $k_T= 1.22\times{10^{ - 5}}$, $k_Q = 1.689\times {10^{ - 7}}$, ${\mathbf{A}}_T= {\rm{diag}}\left( {0.3,0.3,0.25} \right)[kg/s]$, and ${\mathbf{A}}_R = 0.2 \mathbf{I}_3\;[kg/s]$.
The physical constraints correspond to a custom-built single-axis tiltrotor quadrotor in our group with a maximum rotor rotational speed of $10,000[rpm]$ and $-45^\circ \le \beta_i \le 45^\circ$.
We selected the NMPC parameters as $N = 5$, $\mathbf{Q}_v = 5 \times 10^{-4}\mathbf{I}_6$, and $\mathbf{Q}_x = \mathrm{diag}\left(0.04, 0.04, 0.04, 8, 8, 8, 1, 1, 4, 65, 65, 70\right)$.

\subsection*{Scenario 1: Sluggish lemniscate trajectory tracking}

\begin{figure}[t]
    \centering
    \includegraphics[width=\linewidth]{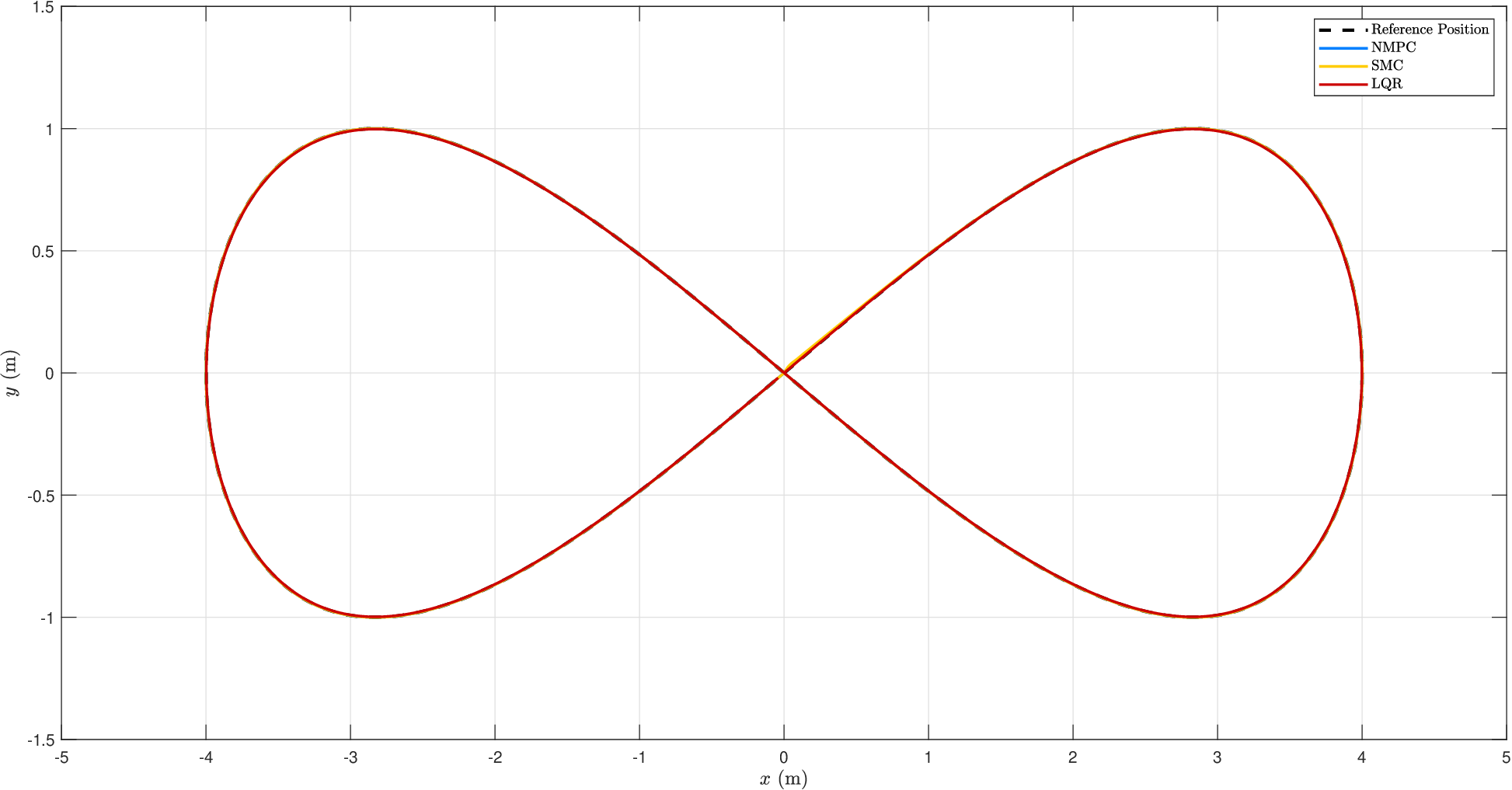}
    \caption{Top view of vehicle position in sluggish trajectory tracking}
    \label{fig:traj1}
\end{figure}

We begin with a straightforward mission: tracking a lemniscate trajectory defined by ${{\boldsymbol{\xi }}_d} = \left[4\sin \left(\frac{{\pi t}}{{20}}\right),\sin \left(\frac{{2\pi t}}{{20}}\right), - 4 \right]^T$ and $\boldsymbol{\eta}_d = \left[ 0, 0, 0 \right]^T$ in the absence of external disturbances (Fig. \ref{fig:traj1}). The maximum acceleration of this trajectory is approximately $0.1[m/{s^2}]$, making it a relatively simple task for all controllers.

\begin{figure}[t]
    \centering
    \includegraphics[trim={0cm 0cm 0cm 0cm}, clip, width = \linewidth]{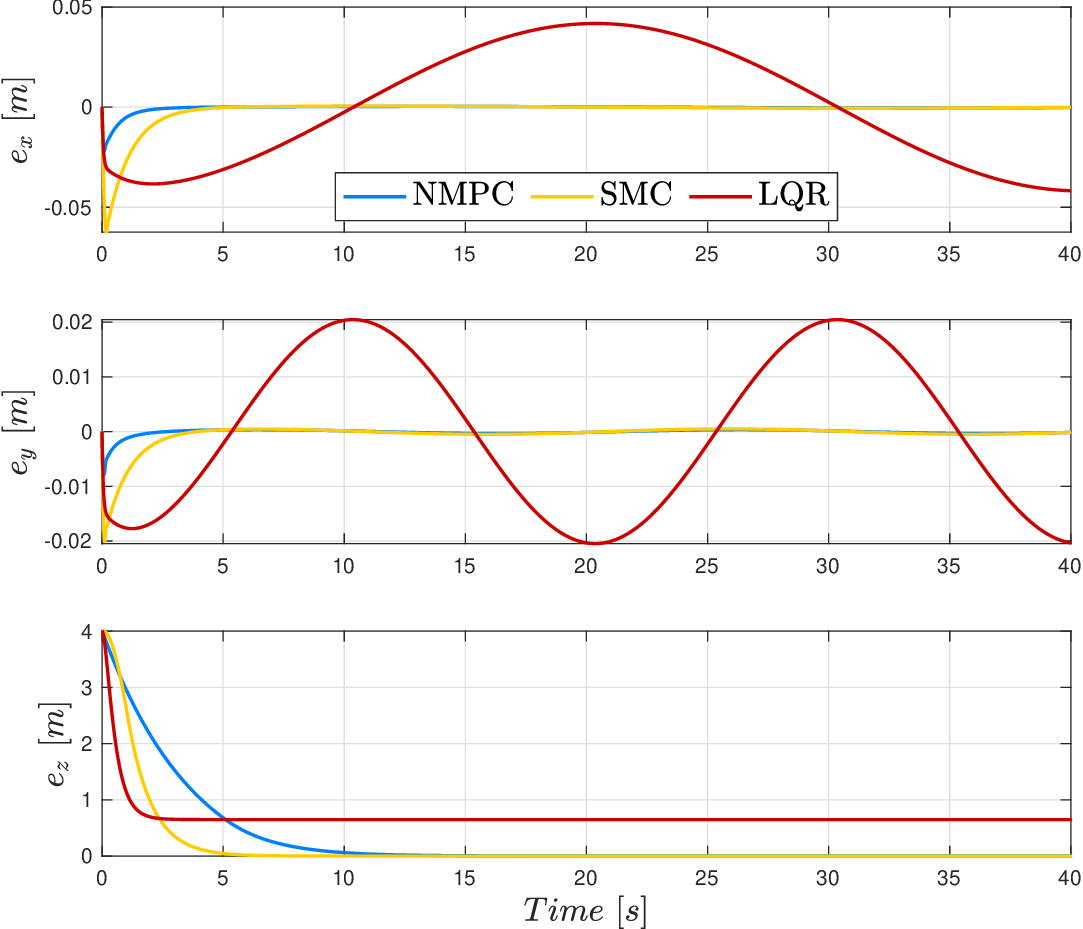}
    \caption{Position error in sluggish trajectory tracking}
    \label{fig:position1}
\end{figure}
\begin{figure}[t]
    \centering
    \includegraphics[trim={0cm 0cm 0cm 0cm}, clip, width = \linewidth]{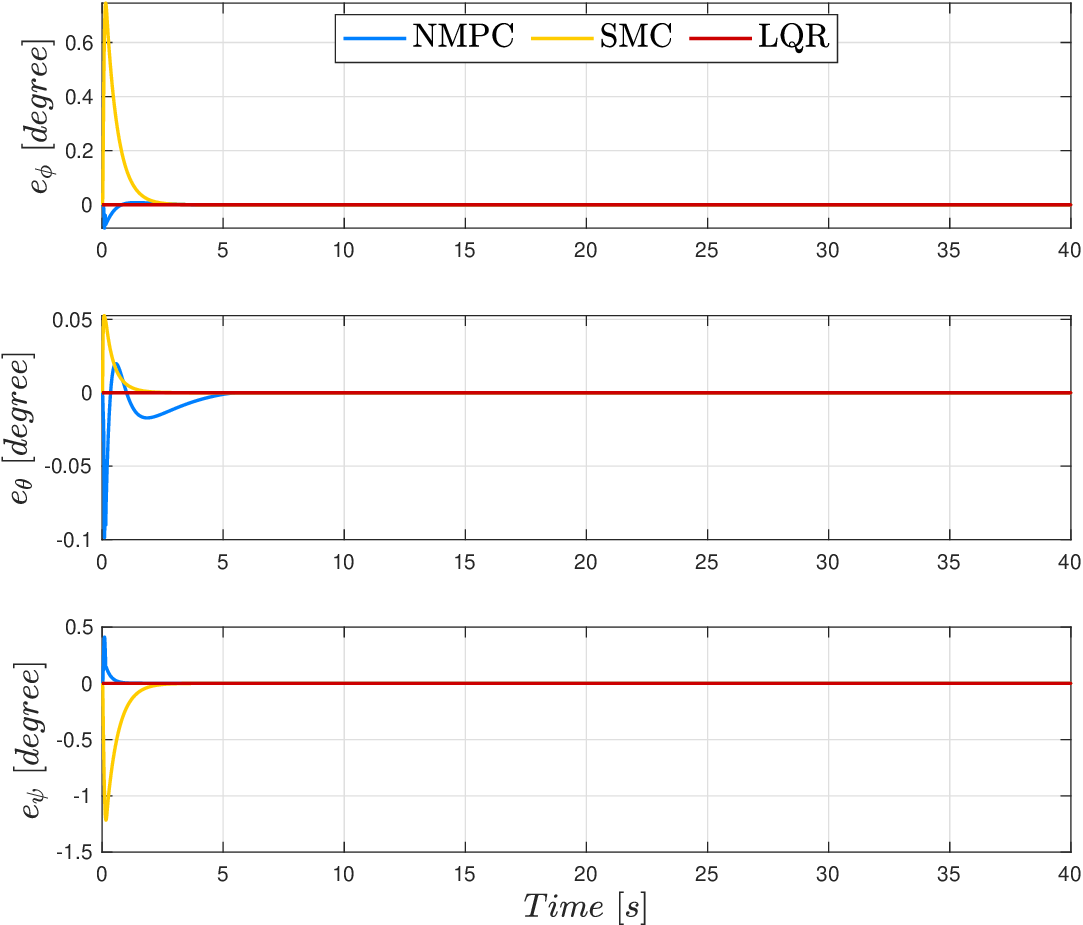}
    \caption{Attitude error in sluggish trajectory tracking}
    \label{fig:attitude1}
\end{figure}
\begin{figure}[t]
   \centering
   \includegraphics[trim={0cm 0cm 0cm 0cm}, clip, width = \linewidth]{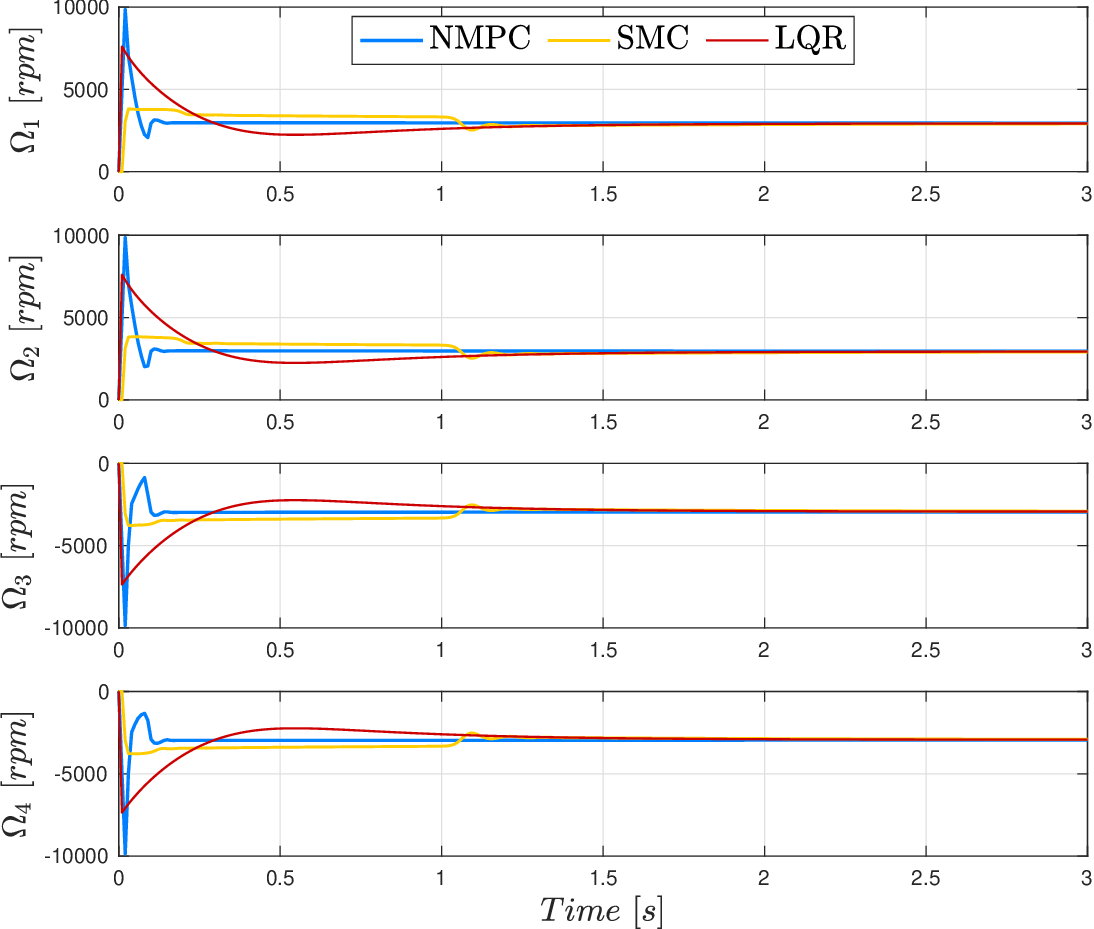}
   \caption{$\Omega_i$ in sluggish trajectory tracking}
   \label{fig:omega1}
\end{figure}
\begin{figure}[t]
   \centering
   \includegraphics[trim={0cm 0cm 0cm 0cm}, clip, width = \linewidth]{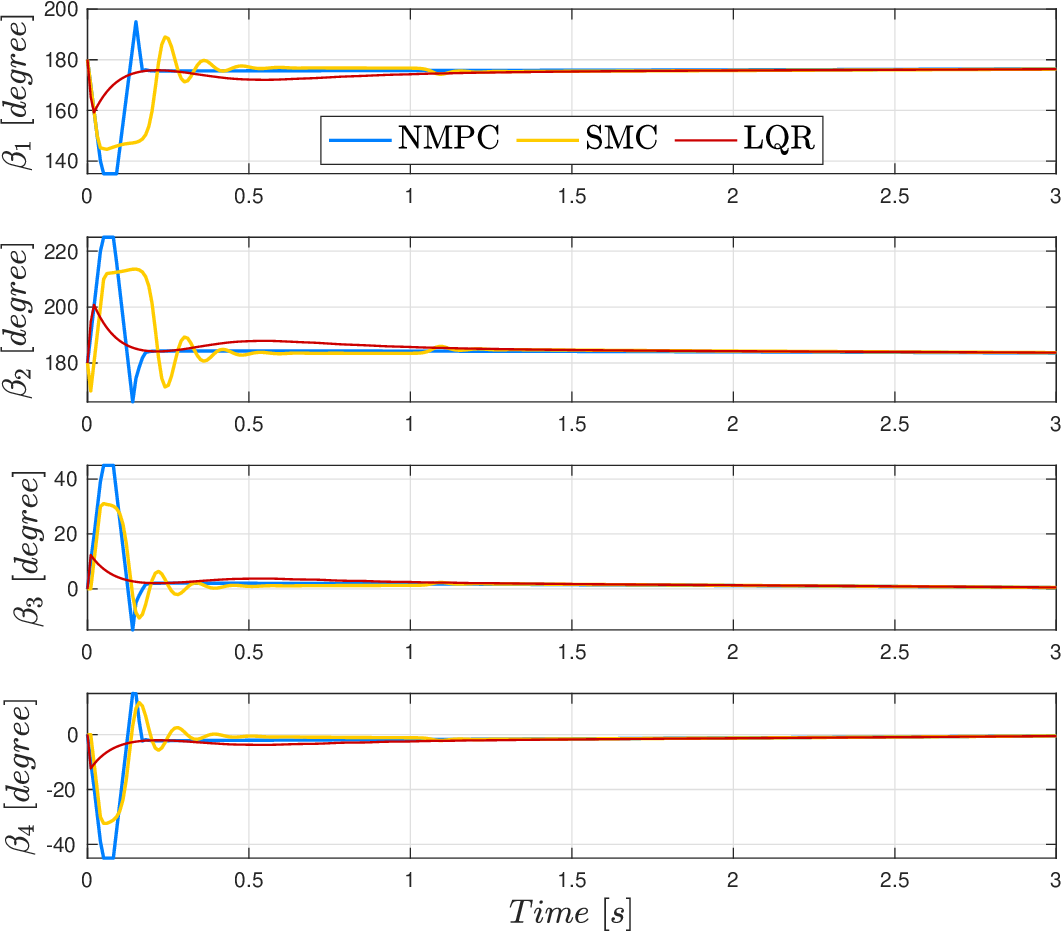}
   \caption{$\beta_i$ in sluggish trajectory tracking}
   \label{fig:beta1}
\end{figure}

Figures \ref{fig:position1} and \ref{fig:attitude1} present the position and attitude tracking errors. All controllers achieve small steady-state errors; however, only NMPC reaches zero steady-state error in all states. Despite our efforts to fine-tune LQR, a steady-state error in the $z$ direction persisted. SMC suffers from a fluctuating error in the $x$ and $y$ directions, and slow convergence to $z_d$ compared to NMPC.

Figures \ref{fig:omega1} and \ref{fig:beta1} present the control inputs during this mission. While different controllers exhibit distinct patterns, all controllers keep the control inputs well within the physical limits of the actuators after a transient state. 
% Note that $0 < \Omega_i$ and $\beta_i$ are bound to.
% , possibly because the desired trajectory is not an agile one.

While NMPC demonstrates superior tracking performance, the differences between the three methods are subtle, primarily because the desired trajectory did not require aggressive maneuvers that would push the vehicle to its actuation limits. The advantages of NMPC become more pronounced as the mission becomes more challenging, such as the one explained below.

\begin{figure}[t]
    \centering
    \includegraphics[width=\linewidth]{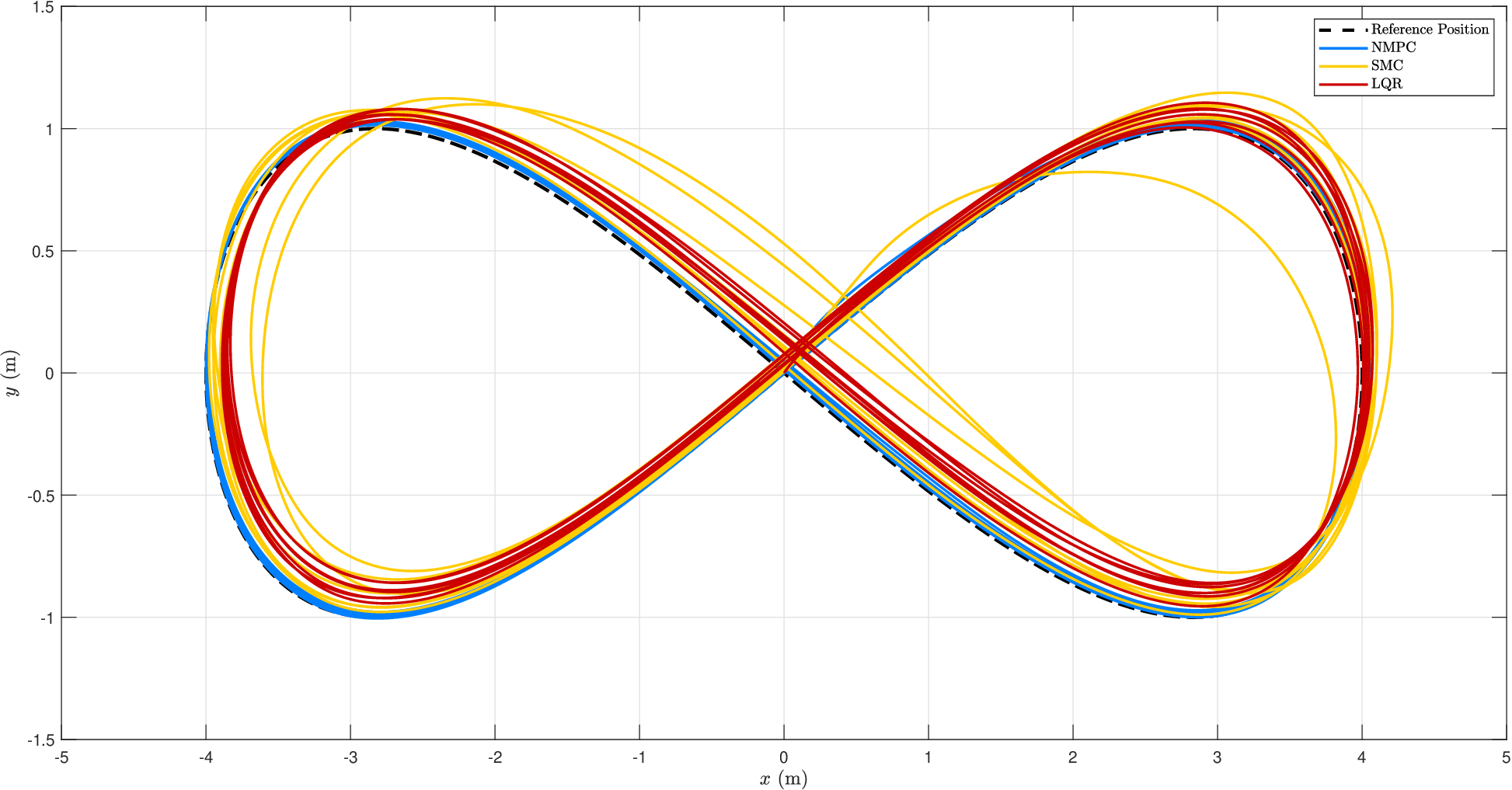}
    \caption{Top view of vehicle position in agile trajectory tracking}
    \label{fig:traj2}
\end{figure}

\begin{figure}[t]
    \centering
    \includegraphics[trim={0cm 0cm 0cm 0cm}, clip, width = \linewidth]{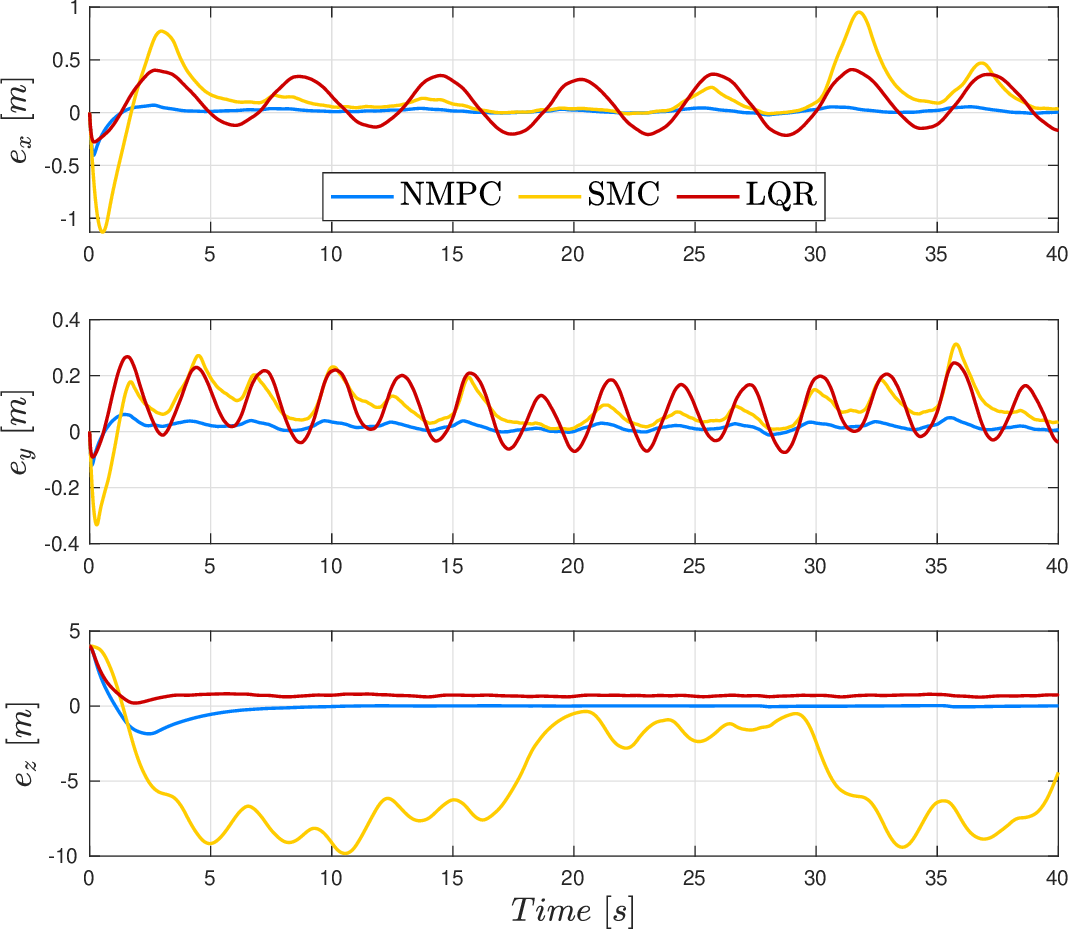}
    \caption{Position error in agile trajectory tracking}
    \label{fig:position2}
\end{figure}
\begin{figure}[t]
    \centering
    \includegraphics[trim={0cm 0cm 0cm 0cm}, clip, width = \linewidth]{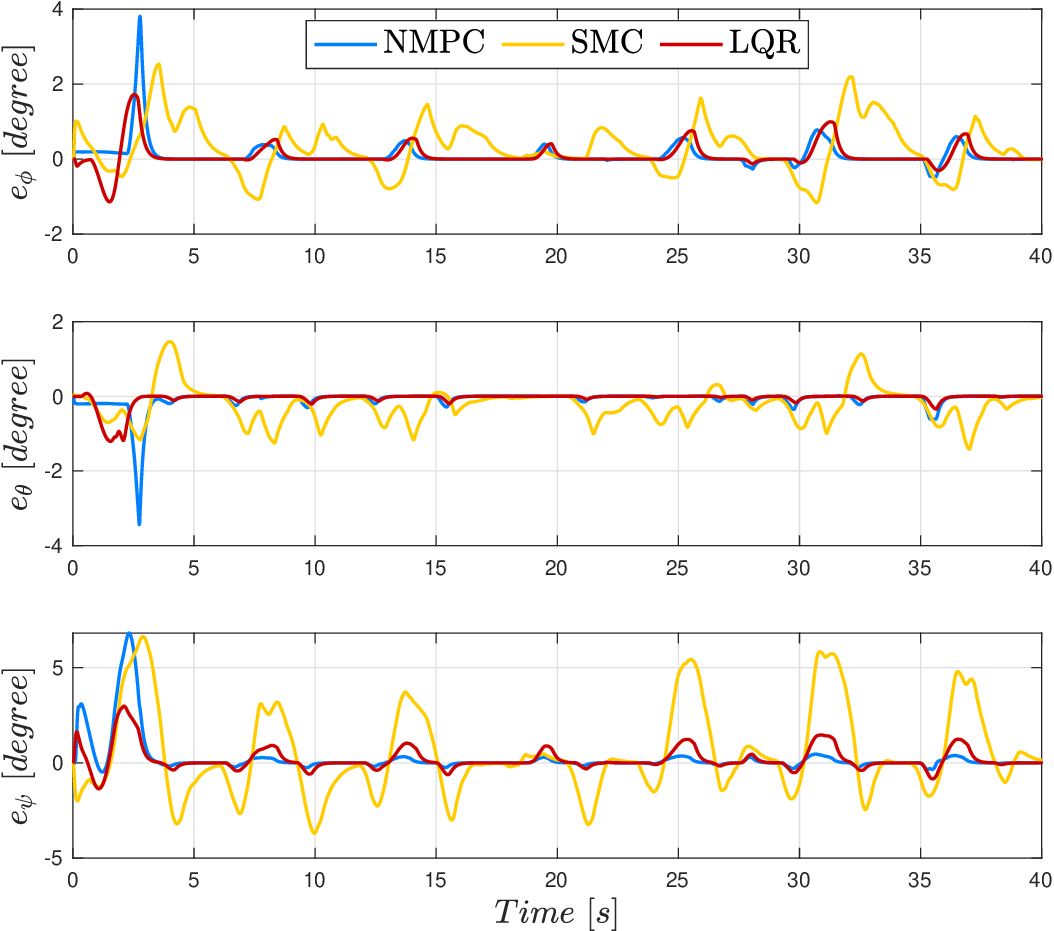}
    \caption{Attitude error in agile trajectory tracking}
    \label{fig:attitude2}
\end{figure}
\begin{figure}[t]
    \centering
    \includegraphics[trim={0cm 0cm 0cm 0cm}, clip, width = \linewidth]{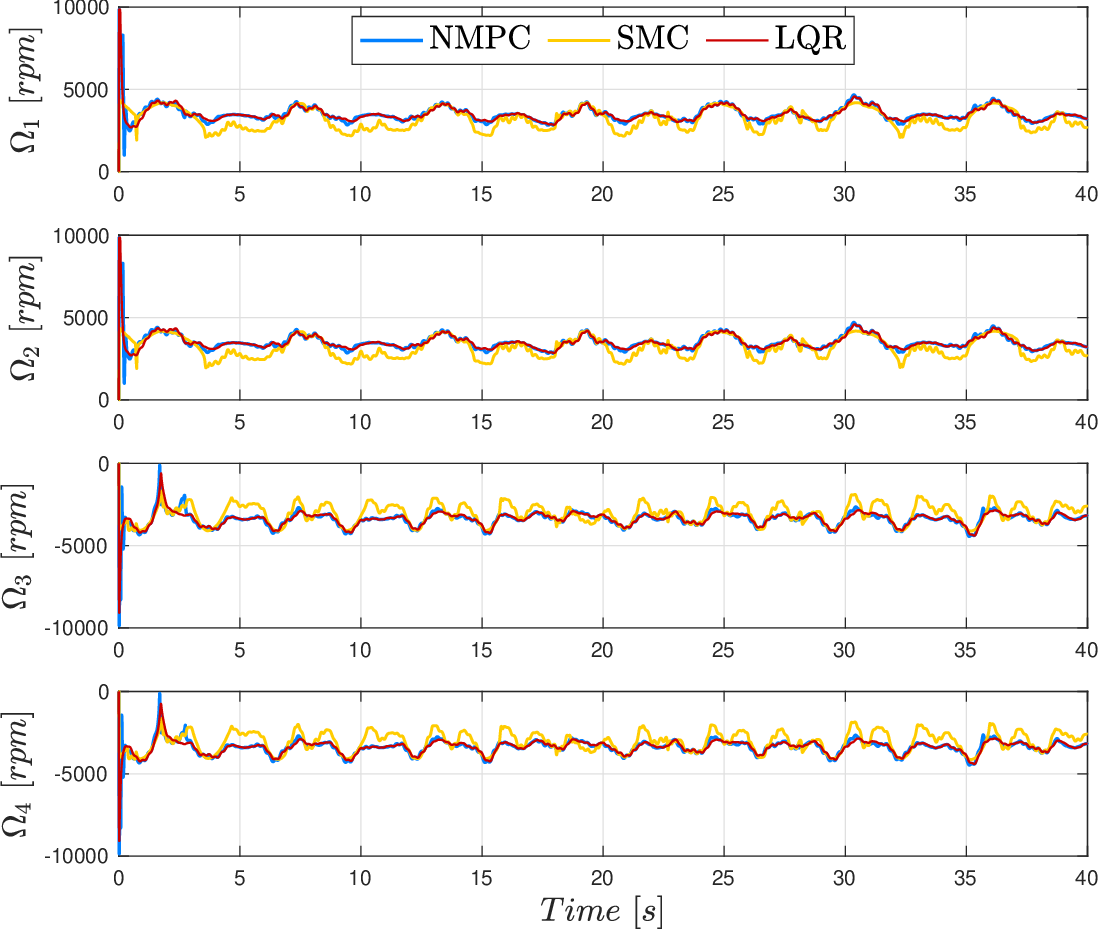}
    \caption{$\Omega_i$ in agile trajectory tracking}
    \label{fig:omega2}
\end{figure}
\begin{figure}[t]
    \centering
    \includegraphics[trim={0cm 0cm 0cm 0cm}, clip, width = \linewidth]{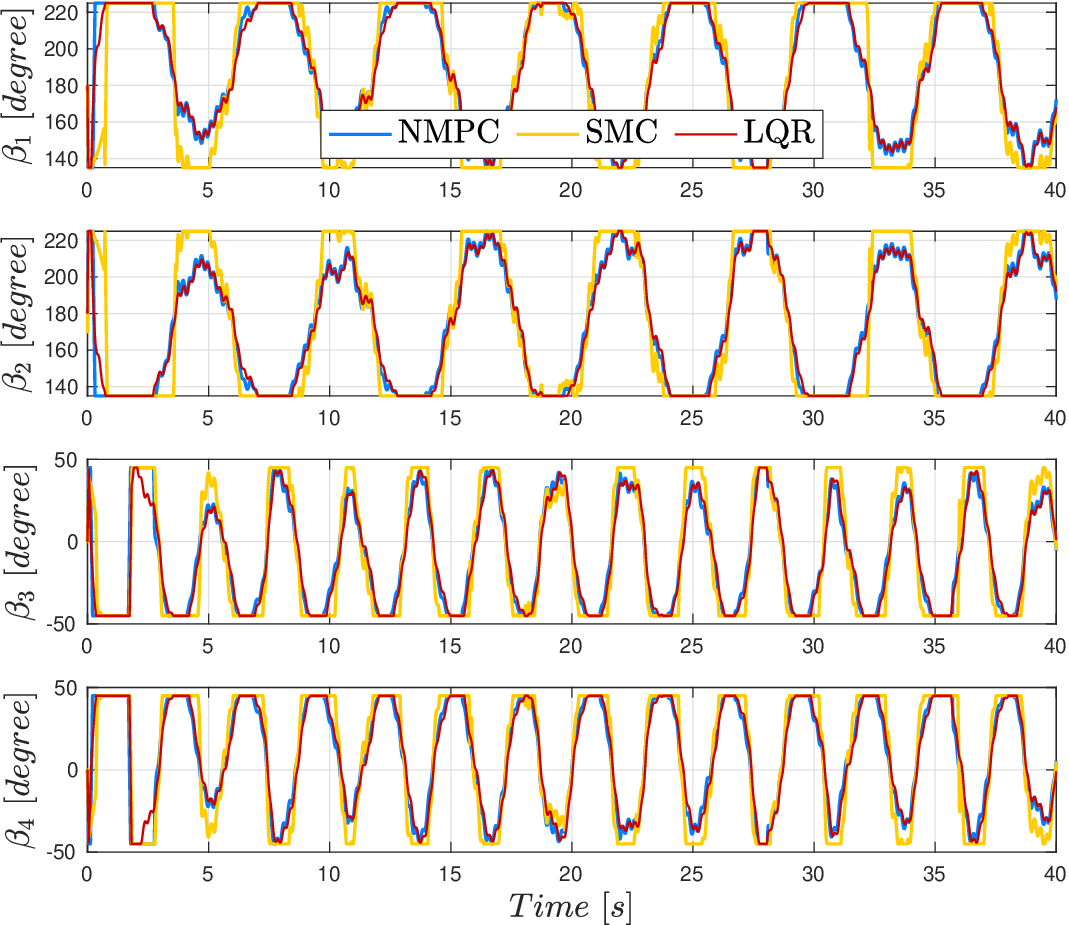}
    \caption{$\beta_i$ agile trajectory tracking}
    \label{fig:beta2}
\end{figure}

\subsection*{Scenario 2: Agile lemniscate trajectory tracking}
We increased the speed of the lemniscate trajectory, reaching to accelerations around $5\;m/{s^2}$, and added a composite sinusoidal signal similar to the one in \cite{izadi2024high} as an external disturbance (Fig. \ref{fig:traj2}). 

Figures \ref{fig:position2} and \ref{fig:attitude2} present the position and attitude tracking errors. Again, NMPC exhibits superior performance, with the lowest magnitude of errors across all states. LQR exhibits significantly larger errors in all states, and a steady-state error in the $z$ direction. SMC performs poorly in the $z$ direction, and its attitude tracking error is also much larger than that of the other controllers.

NMPC's superior performance is attributed to its ability to account for the physical limitations of the actuators when generating virtual control signals. As shown in Figs. \ref{fig:omega2} and \ref{fig:beta2}, the NMPC control inputs exhibit shorter episodes of actuator saturation and less fluctuations in the $\beta_i$ signals, highlighting better control allocation that is achieved by the tight coupling of the controller and allocator, ultimately, leading to improved performance.

\begin{figure}[t]
    \centering
    \includegraphics[trim={0cm 0cm 0cm 0cm}, clip, width = \linewidth]{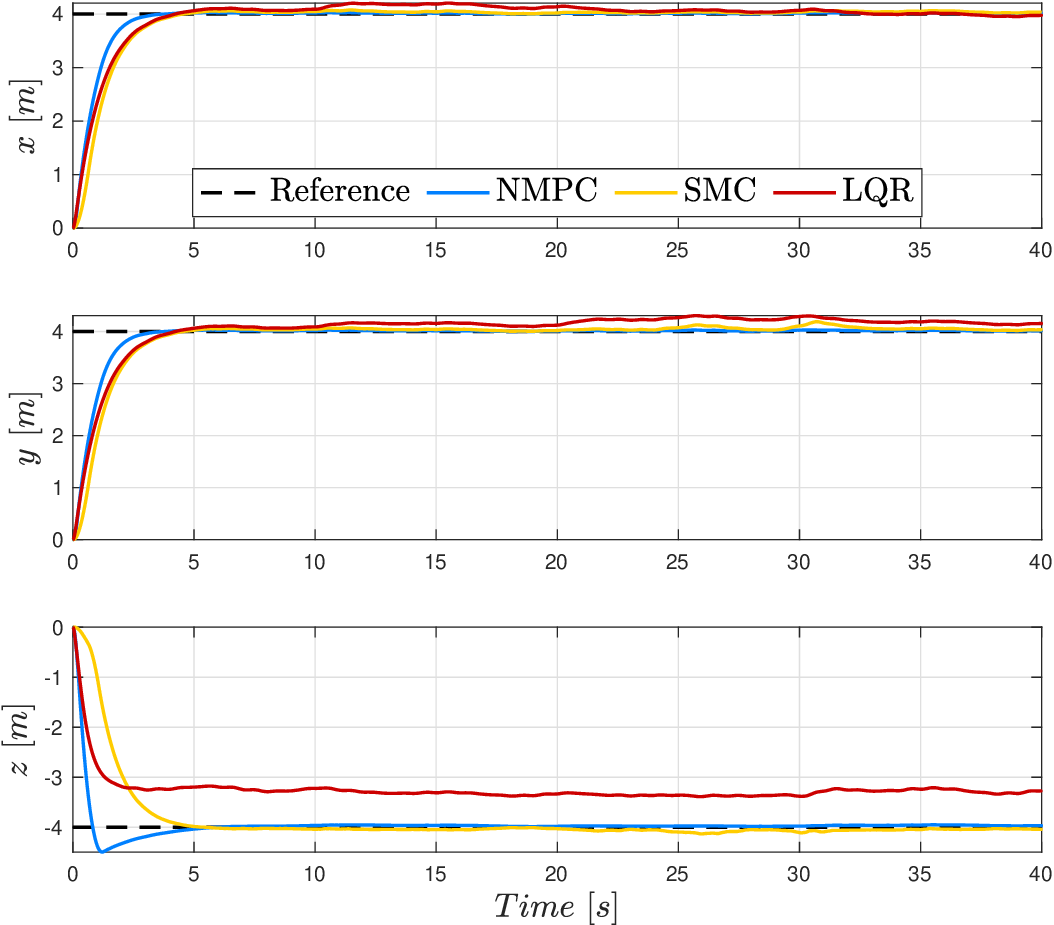}
    \caption{Position in attitude tracking during hover}
    \label{fig:position4}
    \centering
\end{figure}
\begin{figure}[t]
    \includegraphics[trim={0cm 0cm 0cm 0cm}, clip, width = \linewidth]{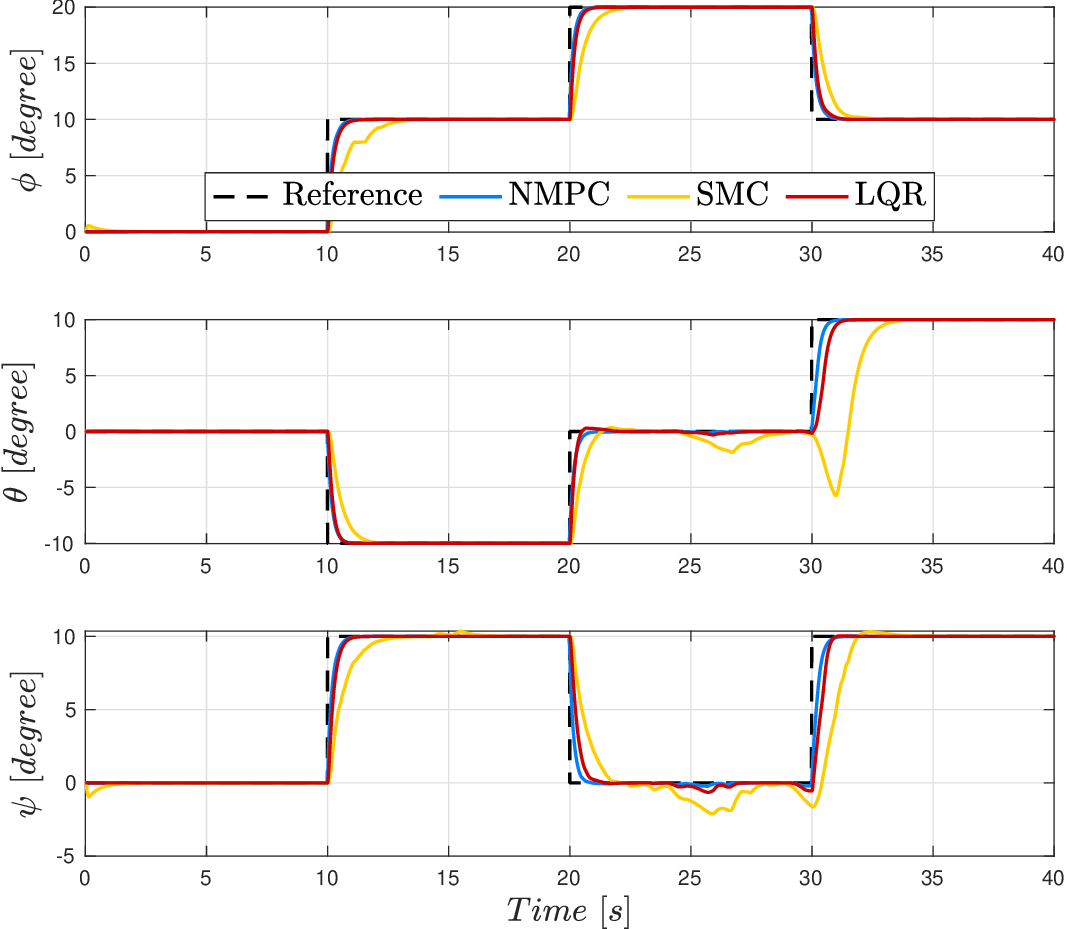}
    \caption{Attitude in attitude tracking during hover}
    \label{fig:attitude4}
\end{figure}
\begin{figure}[t]
    \centering
    \includegraphics[trim={0cm 0cm 0cm 0cm}, clip, width = \linewidth]{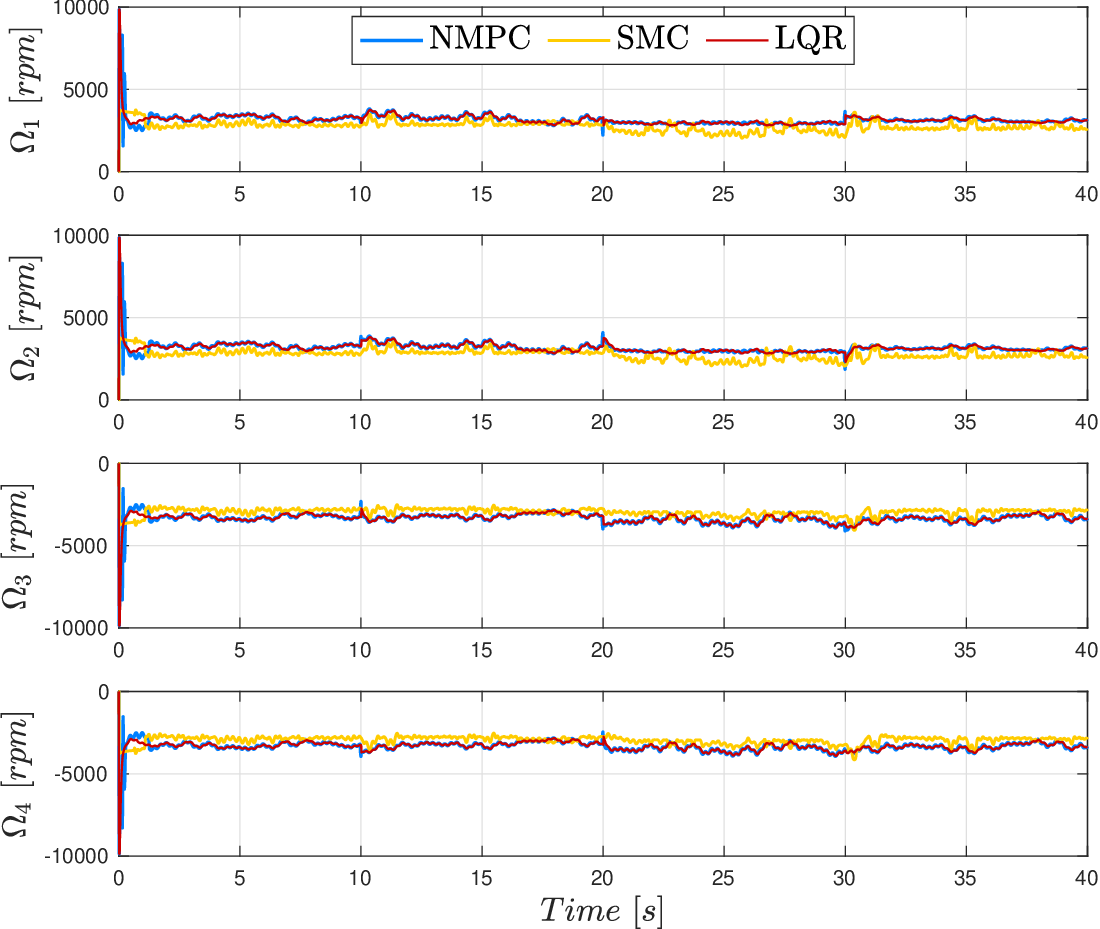}
    \caption{$\Omega_i$ in attitude tracking during hover}
    \label{fig:omega4}
    \centering
\end{figure}
\begin{figure}[t]
    \includegraphics[trim={0cm 0cm 0cm 0cm}, clip, width = \linewidth]{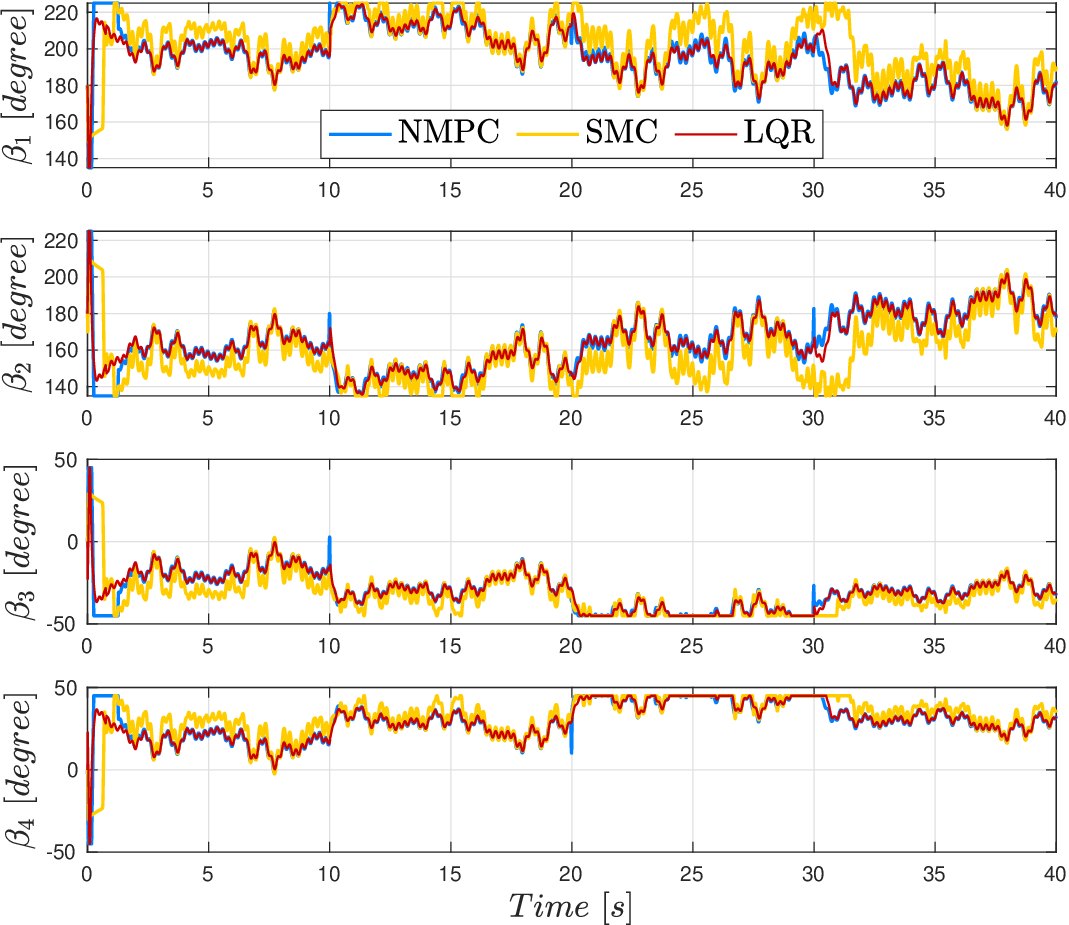}
    \caption{$\beta_i$ in attitude tracking during hover}
    \label{fig:beta4}
\end{figure}
\subsection*{Scenario 3: Attitude tracking in hover}
One advantage of tiltrotor quadrotors over conventional ones is their ability to control pitch and roll while holding a position. We explore the performance of the controller in conducting such maneuvers. The vehicle starts from the origin, flying to $\left[4,4,-4\right]^T$ and holding position despite external disturbances. Then, it tracks different attitude set-points as shown in Figs. \ref{fig:position4}--\ref{fig:beta4}.

Similar to the previous scenarios, NMPC achieves superior position and attitude tracking errors in all states. Interestingly, this is accomplished with smaller control inputs compared to the other controllers. The SMC control input presents undesirable fluctuations that could be attributed to chattering phenomenon.

Overall, these results highlight the superiority of our proposed algorithm for flight control of tiltrotor MRUAVs compared to some of the common existing approaches.

%% file: 05Conclusion.tex
\section{Conclusion}
This paper presented a new flight control framework for tiltrotor MRUAVs. Our approach tightly couples control allocation with the controller and leverages NMPC formulation to effectively handle actuator constraints while maintaining low tracking error. Our comparative study demonstrated the superiority of the proposed method over conventional techniques based on LQR and SMC.
This highlights the viability of our approach for enhanced control precision and robustness, especially in challenging missions.

We utilized the simplest control allocation approach, the pseudo-inverse method, yet our flight control algorithm excelled in all test scenarios. However, the use of more advanced control allocation algorithms could potentially further enhance the algorithm performance. Some advanced control allocation algorithms require solving optimization problems in real-time. Given the computational load of NMPC, integrating advanced control allocation poses a challenge, opening up new research questions on how to effectively implement them in resource-constrained vehicles and conduct hardware experiments. This is an area for future exploration.

Overall, our findings underscore the potential of the proposed approach as a robust and efficient control strategy for tiltrotor MRUAVs, paving the way for more advanced and reliable autonomous flight operations.